\documentclass[manuscript,screen]{acmart}
\AtBeginDocument{%
  }

\usepackage{subcaption}
\usepackage{booktabs}
\usepackage{multirow}
\usepackage{url}

\setcopyright{none}

\acmJournal{TIST}
\acmVolume{}
\acmNumber{}
\acmArticle{}
\acmMonth{0}
\acmYear{}

\begin{document}

\title{StateVLM: A State-Aware Vision-Language Model for Robotic Affordance Reasoning}

\author{Xiaowen Sun}
\email{wenny.xiaowensun@gmail.com}
\orcid{0009-0009-0552-8043}
\correspondingauthor
\affiliation{
  \institution{Department of Informatics, University of Hamburg}
  \city{Hamburg}
  \country{Germany}
}

\author{Matthias Kerzel}
\authornote{Both authors contributed equally to this research.}
\email{matthias.kerzel@uni-hamburg.de}
\affiliation{
  \institution{Department of Informatics, University of Hamburg}
  \city{Hamburg}
  \country{Germany}
}

\author{Mengdi Li}
\authornotemark[1]
\email{li\_mengdi@hotmail.com}
\affiliation{
  \institution{King Abdullah University of Science and Technology}
  \city{Thuwal}
  \country{Saudi Arabia}
}

\author{Xufeng Zhao}
\email{xfz.zhao@gmail.com}
\affiliation{
  \institution{Department of Informatics, University of Hamburg}
  \city{Hamburg}
  \country{Germany}
}

\author{Paul Striker}
\email{paul.jonas950@gmail.com}
\affiliation{
  \institution{Department of Informatics, University of Hamburg}
  \city{Hamburg}
  \country{Germany}
}

\author{Stefan Wermter}
\email{stefan.wermter@uni-hamburg.de}
\affiliation{
  \institution{Department of Informatics, University of Hamburg}
  \city{Hamburg}
  \country{Germany}
}

\renewcommand{\shortauthors}{X. Sun et al.}

\begin{abstract}
Vision-language models have demonstrated strong performance across robotic perception and instruction-following tasks. However, they still struggle with precise spatial reasoning, particularly in predicting object locations and fine-grained object states.
We propose StateVLM, a vision-language model designed to learn fine-grained object representations, including object localization and grasp-relevant region prediction. 
We introduce a joint training objective that integrates an auxiliary regression loss (ARL) with the standard causal language modeling (CLM) objective to improve numerical reasoning and spatial understanding. 
To evaluate whether models can move beyond category-level grounding toward state-aware spatial understanding, we introduce an open-source benchmark, Object State Affordance Reasoning (OSAR), comprising 1,172 scenes with 7,746 individual objects and their corresponding bounding boxes.
Empirical experiments on the Referring Expression Comprehension datasets demonstrate that integrating ARL improves model performance compared with CLM only. Experiments on the OSAR benchmark further demonstrate that StateVLM with ARL achieves an average improvement of 5.2\% over models trained with CLM only. These results show that ARL is particularly beneficial for the complex affordance reasoning tasks that require object state understanding.
The joint training objective in StateVLM demonstrates that integrating an auxiliary regression objective into VLM training improves numerical reasoning, and the OSAR benchmark provides a new testbed for understanding object state affordances in robotics.

\end{abstract}



\maketitle

\section{Introduction}\label{sec:intro}

Vision-language models (VLMs)~\cite{bai2023qwenvl, lin2024sphinx, wang2024visionllm, guo2024regiongpt, zhang2024llavag, chen2023shikra, chen2023minigptv2, pramanick2023jack, you2024ferret, zhang2024ferretv, wang2024cogvlm, wang2025integrated, erol2026unmasking} provide a unified framework that integrates natural language understanding~\cite{ichter2023saycan, pope2025evaluating}, visual perception~\cite{hu2024look, radford2021learning}, and commonsense reasoning~\cite{ren2023robots} for robotics applications.
However, when applied to robotic settings, VLMs remain challenged by tasks that require precise spatial reasoning and numerical outputs, such as object localization and spatial grounding~\cite{li2024spatial, chen2024spatialvlm, kamath2023s}.

Most modern VLM architectures adopt large language models (LLMs)~\cite{team2024qwen2, touvron2023llama} as their backbone and integrate visual perception modules~\cite{dosovitskiyimage, radford2021learning} within them. Since LLMs are originally trained for sequence prediction, many VLMs inherit a sequence-based formulation for object detection and localization. For instance, Pix2Seq~\cite{chen2022pixseq} predicts object bounding boxes and class labels as discrete token sequences. Similarly, state-of-the-art VLMs such as SPHINX~\cite{lin2024sphinx}, Shikra~\cite{chen2023shikra}, and LLAVA-G~\cite{zhang2024llavag} adopt this paradigm.
This Pix2Seq formulation models continuous spatial coordinates as discrete token generation. However, tokenized coordinates may weaken geometric structure, since spatial continuity is not explicitly preserved under discrete tokenization, unlike regression-based objectives. We therefore hypothesize that applying a purely sequence-based formulation to numerical tasks may lead to inefficient learning of spatially precise representations.

To address this limitation of tokenized continuous coordinates as sequences, NExT-Chat~\cite{zhang2023nextchat} explores object detection from a regression perspective using a Pix2Emb framework. It introduces a box encoder and decoder to represent continuous coordinates as embeddings. The key difference between Pix2Seq and Pix2Emb lies in their output representations: Pix2Seq generates discrete sequences, whereas Pix2Emb combines sequence generation with continuous embedding prediction. However, despite using similar training data for detection, NExT-Chat performs slightly worse than Shikra-7B~\cite{chen2023shikra} on referring expression comprehension (REC).
Zhang et al. hypothesize that this gap arises from two factors. First, balancing sequence and embedding losses is non-trivial, while Pix2Seq avoids this issue by optimizing only sequence loss. Second, LLMs are not explicitly pre-trained for regression-style objectives, which may further complicate optimization.

\begin{figure*}[t]
\centering
\includegraphics[width=0.8\textwidth]{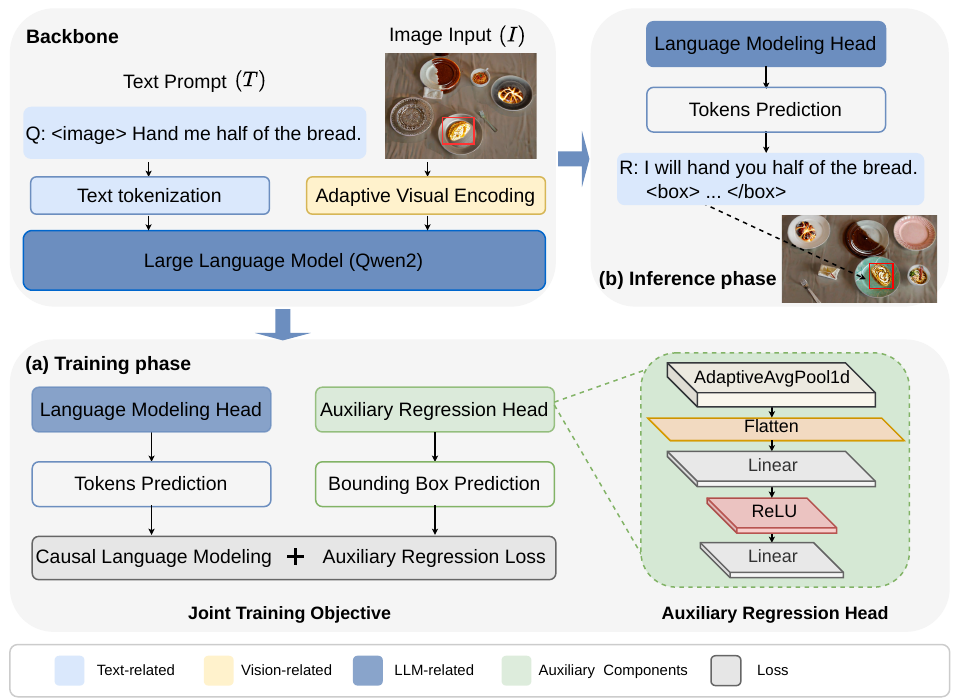}
\caption{Schematic overview of \textbf{StateVLM} architecture and its operation during training and inference. During the \textbf{training phase (a)}, a specially designed auxiliary regression head converts sequence predictions into embedding predictions for the computation of an auxiliary regression loss, which is jointly optimized with causal language modeling. During the \textbf{inference phase (b)}, however, the model continues to rely on the language modeling head to generate sequences composed of text and bounding boxes, since the LLM backbone of the VLM is originally trained for sequence prediction.}
\label{fig:method}
\end{figure*}

Motivated by these research gaps, we propose a less intrusive alternative to Pix2Emb methods. Our approach introduces an auxiliary regression head trained with an auxiliary regression loss (ARL), while keeping the standard Pix2Seq sequence-generation interface unchanged at inference time. This design preserves the original autoregressive generation paradigm while leveraging regression signals during training, thereby mitigating optimization difficulties associated with explicit embedding-based decoding, as shown in Fig.~\ref{fig:method}.

Importantly, improving object localization alone is not sufficient to evaluate whether models can perform state-level affordance reasoning, which requires fine-grained understanding of object states and their spatially relevant regions. Existing object referring expression benchmarks such as RefCOCO, RefCOCO+, and RefCOCOg~\cite{kazemzadeh2014referitgame, mao2016generation} focus on category-level grounding and spatial identification rather than object states. Meanwhile, existing benchmarks on object states primarily focus on recognition and classification rather than localization~\cite{gouidis2024exploring, jelodar2018identifying, spisak2023clarifying, gouidis2022detecting}, and only a few datasets consider object state changes~\cite{li2024skt, nguyen2024oscar}.
Even with recent advances in modeling object states, it remains unclear whether existing methods support state-level affordance reasoning. Affordance reasoning at the object state-level~\cite{liu2018eccv, gibson1977affordances} is essential for robotic manipulation, where actions depend on both object identity and state-specific spatial regions. Unlike object categories, object states are dynamic and can vary across instances and over time~\cite{spisak2023clarifying,grauman2022ego4d,yang2024attribute}. For example, two plates labeled as ``dirty plate'' may require different grasping regions depending on the spatial distribution of the dirt.

To address the lack of object state understanding benchmarks for robotic manipulation, we introduce an open-source benchmark for object state localization and affordance reasoning, designed to evaluate whether models can move beyond category-level grounding toward fine-grained state-aware spatial understanding.

Our primary contributions are summarized as follows:

\begin{enumerate}

\item \textbf{StateVLM.} We propose StateVLM\footnote{\url{https://github.com/Xiao-wen-Sun/StateVLM_V0.git}}, a vision-language model designed to learn fine-grained object representations, including object localization, object state grounding, and grasp-relevant region prediction.

\item \textbf{Auxiliary Regression Learning (ARL).} We introduce a joint training objective that integrates an auxiliary regression loss (ARL) with the standard causal language modeling objective. 

\item \textbf{OSAR benchmark.} We introduce Object State Affordance Reasoning (OSAR), an open-source benchmark for evaluating whether models can move beyond category-level grounding toward state-aware spatial understanding. OSAR comprises 1,172 scenes with 7,746 annotated objects and their corresponding bounding boxes.

\item \textbf{Empirical evaluation.} We evaluate StateVLM on RefCOCO, RefCOCO+, and RefCOCOg~\cite{kazemzadeh2014referitgame, mao2016generation}. Experiments show that ARL improves fine-tuning stability and localization performance. We further provide baseline results on OSAR using StateVLM, establishing initial benchmarks for state-aware affordance reasoning.

\end{enumerate}

The remainder of this paper is structured as follows: 
Section~\ref{sec:related-work} reviews existing approaches to tackle object state reasoning and points out the limitations of current methods.
Section~\ref{sec:dataset-affordance-reasoning} describes the task definition and provides a summary of our proposed benchmark, named OSAR. 
The architecture and training procedure of StateVLM are detailed in Section~\ref{sec:statevlm}.
Section~\ref{sec:experiments} presents and analyzes the experimental results, then discusses our findings, highlighting the role of the auxiliary regression loss in improving performance. 
Finally, Section~\ref{sec:conclusion} concludes the paper and suggests directions for future work.

\section{Related Work}\label{sec:related-work}

\subsection{VLMs for Referring Expression Comprehension}

Referring Expression Comprehension (REC) is a fundamental region-level image understanding task that seeks to identify and localize a target object described by a natural language expression within a given scene~\cite{kazemzadeh2014referitgame,mao2016generation}. 
In recent years, State-of-the-art VLMs have achieved remarkable performance on this task~\cite{lin2024sphinx, chen2023shikra, chen2023minigptv2, bai2023qwenvl, zhang2024llavag, pramanick2023jack, you2024ferret, zhang2024ferretv, wang2024cogvlm}, significantly surpassing traditional non-VLM approaches~\cite{wang2022ofa, deng2021transvg, chen2020uniter, gan2020villa, yang2022unitab, liu2023gdino}. 
This improvement can largely be attributed to the integration of visual representations or embeddings into LLMs.
Due to their training on massive and diverse text corpora, LLMs exhibit strong linguistic interpretation abilities and extensive commonsense reasoning, both of which are crucial for advancing embodied AI and robotic perception.
Nevertheless, these models demand substantial computational resources and prolonged multi-stage training processes, which present notable practical challenges.

The computational resources and training times of the most representative VLMs for the REC task are different. 
Depending on the number of transformer layers in the LLM and the scale of the visual backbone, they typically contain either 7B(illion) or 13B parameters.
Furthermore, the models adopt several different training configurations. 
First, they are trained on computational clusters of varying sizes, such as 8, 32, or 256$\times$A100 GPUs.
Second, they employ various training strategies, including one-stage, two-stage, or multi-stage training. 
Third, training durations are reported inconsistently, with some measured in wall-clock time and others in training steps.

Therefore, it is difficult to draw a rigorous scientific conclusion because we lack controlled variables~\cite{wohlin2012experimentation}. It remains unclear whether simply combining two distinct data distributions (discrete text and continuous location data) results in inefficient training for bounding box coordinate prediction. 
To test this hypothesis, we introduce a joint training objective that combines an auxiliary regression loss (ARL) with the causal language modeling (CLM) objective and evaluate it against a CLM-only fine-tuning baseline on adapted REC benchmarks.

\subsection{Affordance Reasoning: From Object Categories to Properties and States}

Object affordance refers to the range of actions an agent can perform with an object, as perceived through its properties~\cite{gibson1977affordances, norman1988psychology}. 
A robust understanding of object categories (e.g., `a cup', `a plate', and `an apple') provides a foundational basis for such reasoning. 
Building on the object categories, a model can further infer an object's potential affordances by leveraging fine-grained object properties.

Object attributes, as generalizable properties, are central to this reasoning process. For instance, instead of relying solely on category recognition, Yang et al.~\cite{yang2024attribute} developed a robotic grasping method based directly on object attributes. Similarly, Attr-POMDP~\cite{yang2022interactive} presents an attribute-guided formulation of a partially observable Markov decision process for task disambiguation. 
In the Octopi system~\cite{yu2024octopi}, the authors selected hardness, roughness, and bumpiness as key physical attributes for physical reasoning.
Another prominent direction in affordance reasoning, particularly for robotic manipulation, involves physically grounded methods~\cite{huang2024avlm, li2024manipllm, huang2024manipvqa}.
PhyGrasp~\cite{guo2024phygrasp} is designed to accurately assess the physical properties of object parts to determine optimal grasping poses.

These fine-grained object properties lack a unified representation, and both their values and the number of properties may vary across objects.
We ground the concept of an object state in object-oriented programming~\cite{booch2008object}, where it is defined as follows: ``The state of an object encompasses all of the (usually static) properties of the object, plus the current (usually dynamic) values of each of these properties.''
In our context, the properties refer to object properties (usually static), such as color, shape, size, weight, texture, cleanliness, and rigidity. The values of these properties are dynamic, meaning they can change over time or remain constant, and different properties may change at different rates.
Our method centers on understanding an object's state and reasoning about its affordances, which has been less explored so far.

\subsection{The Missing Link Between Object State and Manipulation}

Previous benchmarks have primarily focused on object state recognition and classification~\cite{gouidis2024exploring}.
For example, the cooking state recognition challenge~\cite{jelodar2018identifying} focused on 18 types of objects and their corresponding states, such as diced, grated, or creamy paste.
Another dataset~\cite{spisak2023clarifying} classified whether a container was full or half empty.
Furthermore, the Object State Detection Dataset~\cite{gouidis2022detecting} includes a larger number of objects and state classes.
Another set of datasets focuses on detecting or anticipating object state changes.
For example, a subset of Ego4D~\cite{grauman2022ego4d} addresses object state change classification and detection.
Ego4D-OSCA~\cite{manousaki2024anticipating} is a dataset for anticipating object state changes from Ego4D video sequences.
The OSCaR dataset~\cite{nguyen2024oscar} focuses on object state captioning and state change representation. 
State-aware Keypoint Trajectories (SKT)~\cite{li2024skt} provides a synthetic dataset encompassing a broad spectrum of garment configurations, ranging from flat to deformed and folded states.
However, none of these datasets explicitly focus on how an object's state influences its manipulation.

Overall, further investigation is needed to determine whether simply combining discrete and continuous data distributions leads to inefficient training for bounding-box coordinate prediction in VLMs.
Designing a multimodal dataset is essential for analyzing this hypothesis at the object state-level, especially for affordance-based reasoning for object manipulation.

\section{Multimodal Dataset for Object State Affordance Reasoning}
\label{sec:dataset-affordance-reasoning}

\begin{figure*}[h]
\centering
\includegraphics[width=0.7\textwidth]{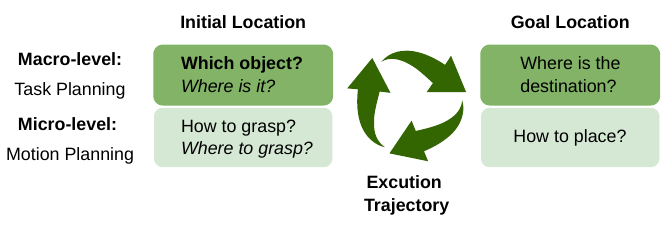}
\caption{Decomposition of robotic manipulation into macro- and micro-level tasks.
The macro-level task involves task planning, identifying objects, and their destinations.
The micro-level task involves motion planning, grasping, and placing the objects.
While an object's state can affect all steps, we focus on its impact on localization, specifically determining the target object, its position, and the grasping location.}
\label{fig:object-manipulation}
\end{figure*}

Robotic manipulation can be decomposed into macro- and micro-level tasks.
The macro-level task involves task planning, where the robot identifies objects and reasons about their destinations.
The micro-level task involves motion planning for grasping and placing the object, as shown in Fig.~\ref{fig:object-manipulation}.
Our previous study, OSSA~\cite{sun2024state}, focused on the macro-level task of robotic task planning. However, this study did not address object state localization and affordance reasoning, both of which are crucial for robotic manipulation. To investigate this missing aspect, we propose and make available a novel benchmark, \textbf{Object State Affordance Reasoning (OSAR)}.

\subsection{Task Definition}

\paragraph{Object Detection: Object State Localization}

Classic referring expression comprehension benchmarks, including RefCOCO, RefCOCO+, RefCOCOg~\cite{kazemzadeh2014referitgame, mao2016generation}, and Flickr30K~\cite{hodosh2013flickr}, focus on object category-level understanding, spatial identification, and grounding.
However, they do not explicitly address object state expressions used to refer to object locations, which are crucial for robotic manipulation.
To address this gap, we construct a new referring expression comprehension dataset for object state localization.
The expressions provide explicit descriptions of the target object with respect to its state and spatial location. The dialog format is shown in Table~\ref{tab:list-of-conversations}, where the expressions are instantiated using object states present in the corresponding scenes, such as `empty plate', `dirty plate', and `bowl with noodles', as illustrated in Table~\ref{tab: ambiguous_opposites} and Fig.~\ref{fig:heatmap}.

\paragraph{Affordance Reasoning: Object Grasp Prediction}

In contrast to object state localization, object grasp prediction requires consideration of physical common sense. For instance, in everyday kitchen scenarios, references are typically made to the object itself rather than to its container. During physical interaction, the decision of whether to grasp a container or its contents depends on the properties and state of the contained item. Specifically, one would grasp a container (e.g., a bowl) to pass noodles, whereas passing an apple does not require a container.
Accordingly, we introduce the second task, termed object grasp prediction, whose format is identical to that of object detection (see Table~\ref{tab:list-of-conversations} for affordance reasoning), except that the predicted bounding box corresponds to the region intended for grasping.

\begin{table*}[t]
    \centering
    \caption{Dialog templates in OSAR. The placeholder [target] is instantiated with the state of objects present in the corresponding scenes, such as `empty plate', `dirty plate', or `bowl with noodles'. The bounding box coordinates $\langle \text{box} \rangle x_1, y_1, x_2, y_2 \langle \text{/box} \rangle$ denote the object location in the object detection task and the grasp location in the affordance reasoning task, where $(x_1, y_1)$ corresponds to the top-left corner and $(x_2, y_2)$ corresponds to the bottom-right corner.}
    \vspace{1em}
    \resizebox{0.8\textwidth}{!}{
    \begin{tabular}{c | c | l}
    \toprule
    \textbf{Task} & \multicolumn{2}{c}{\textbf{Conversations}} \\ 
    \cmidrule(lr){2-3}

    \textbf{Types} & \textbf{Roles} & \textbf{Content} \\  
    \midrule
   
                        & User: & ``Show me the [target].'' \\
      \textbf{Object}   & StateVLM: & ``response: Here is the [target]. $\langle \text{box} \rangle x_1, y_1, x_2, y_2 \langle \text{/box} \rangle$.'' \\
    \textbf{detection}  & User: & ``Where is the [target]?'' \\
    (object state & StateVLM: & ``response: Here is the [target]. $\langle \text{box} \rangle x_1, y_1, x_2, y_2 \langle \text{/box} \rangle$.'' \\
  localization) & User: & ``Where is the location of the [target]?'' \\
 & StateVLM: & ``response: Here is the location of the [target]. $\langle \text{box} \rangle x_1, y_1, x_2, y_2 \langle \text{/box} \rangle$.'' \\
    \midrule
   
            & User: & ``Hand me the [target].'' \\
    \textbf{Affordance} & StateVLM: & ``response: Sure. I will hand you the [target]. $\langle \text{box} \rangle x_1, y_1, x_2, y_2 \langle \text{/box} \rangle$.'' \\
    \textbf{reasoning} & User: & ``Pass me the [target].'' \\
    (object grasp & StateVLM: & ``response: Alright, I will pick up the [target] for you. $\langle \text{box} \rangle x_1, y_1, x_2, y_2 \langle \text{/box} \rangle$.'' \\
    prediction) & User: & ``Give me the [target].'' \\
     & StateVLM: & ``response: Okay, I will give you the [target]. $\langle \text{box} \rangle x_1, y_1, x_2, y_2 \langle \text{/box} \rangle$.'' \\
    \bottomrule
    \end{tabular}
    }
    \label{tab:list-of-conversations}
\end{table*}

\begin{table*}[t]
    \centering
    \caption{Ambiguous opposites and synonyms in OSAR: Tableware states reflect the shifting semantics of use and cleanliness.}
    \vspace{1em}
    \resizebox{0.75\textwidth}{!}{
    \begin{tabular}{c l l}
    \toprule
    \textbf{Object Category} & \textbf{Physical Condition} & \textbf{Semantic Interpretation} \\ 
    \midrule
    \multirow{2}{*}{\textbf{Plate, Bowl}} & Empty, no residue  & clean / unused  \\
     \multirow{2}{*}{\textbf{Cup, Mug, Glass}} &Empty with little residue & used / dirty\\
                                          & With / holding contents & used / dirty\\
    \midrule
    \multirow{2}{*}{\textbf{Bottle}} & Cap open & possibly used (depends on liquid level) \\
                                     & Cap closed / unopened & possibly unused (depends on liquid level)\\
    \midrule
    \multirow{2}{*}{\textbf{Spoon, Fork, Knife}} & No residue & clean / unused \\  
                                      & Has residue & dirty / used \\
    \midrule
    \multirow{4}{*}{\textbf{Napkin}} & Has residue &\multirow{2}{*}{used / dirty} \\               
                                     & Unfolded & \\
                                     \cmidrule(lr){2-3} 
                                     & No residue & \multirow{2}{*}{clean / unused} \\  
                                     & Folded neatly &  \\
    \bottomrule
    \end{tabular}
    }
    \label{tab: ambiguous_opposites}
\end{table*}

\subsection{Benchmark Dataset}
\label{sec:benchmark-dataset}

\subsubsection{Visual Scenes}

\begin{figure*}[h]
\centering
\includegraphics[width=1\textwidth]{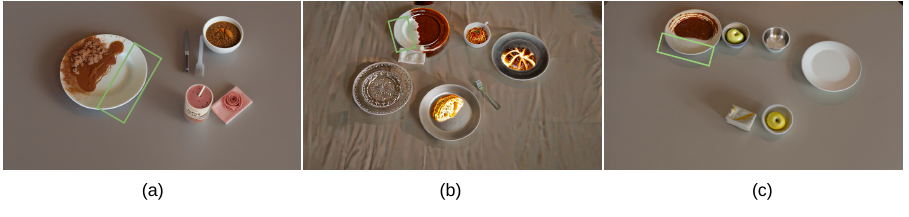}
\caption{Example scenes in OSAR: (a) \textbf{Simple scenes} are defined as those with only one object from each category.
(b) and (c) \textbf{Complex scenes} are defined as those with multiple objects from each category in various states. The green box marks the ideal grasp region; this is an example of how an object’s state affects its affordance, as the area covered with food should be avoided when grasping.}
\label{fig:scenes}
\end{figure*}

\begin{figure*}[h]
\centering
\includegraphics[width=0.9\textwidth]{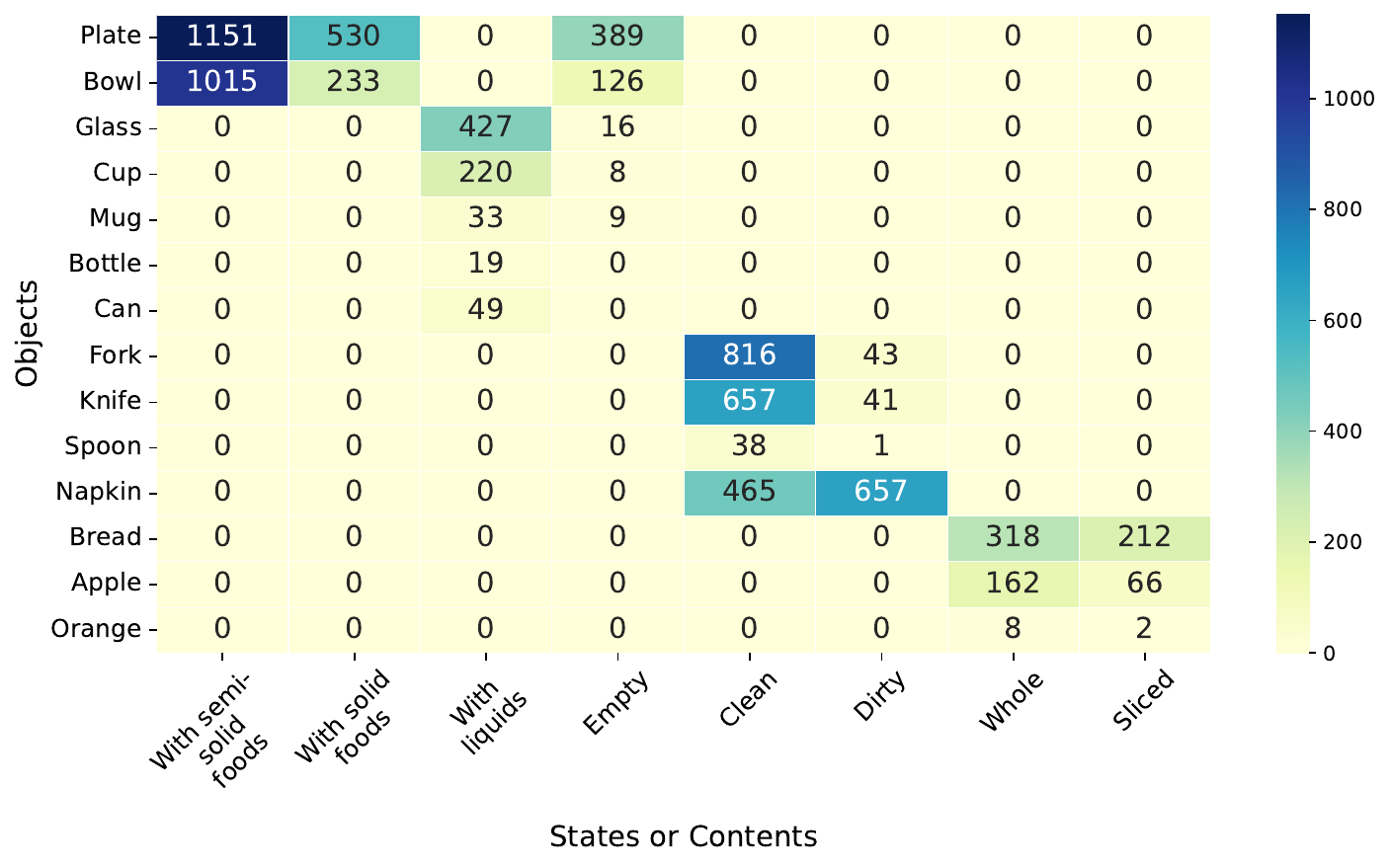}
\caption{Object statistics in OSAR. Semi-solid foods usually include sauces, pasta, soup, noodles, and creams. Solid foods typically refer to various states of apples, bread, and oranges. Liquids encompass a variety of drinks, beverages, and juices.}
\label{fig:heatmap}
\end{figure*}

To reduce the cost of collecting and annotating real-world data, we render synthetic scenes in Blender\footnote{https://www.blender.org/} and subsequently employ the Stable Diffusion model~\cite{rombach2022stable} to generate the final images.
In total, we selected 1172 images based on quality, comprising 780, 267, and 125 images for templates (a), (b), and (c), respectively, as illustrated in Fig.~\ref{fig:scenes}. From each scene, we selected 11.2\% of the images as test samples.
The final dataset comprises 7746 individual objects, of which 11.17\% belong to the test set. These objects were in different states within their respective categories, as shown in Fig.~\ref{fig:heatmap}. 

\subsubsection{Annotation Rules}
When assessing the state of kitchen tools, multiple dimensions must be considered, including contents, usage, hygiene, physical condition, position, accessibility, and operational status.
OSAR specifically focuses on \texttt{clearing the table} after a meal. Our scope within this task is defined as follows: containers are analyzed based on their contents, specifically whether they are empty or contain food, and what type of food they contain.
Cutlery and napkins are evaluated based on hygiene and readiness for usage; they are categorized simply as clean or dirty.
For object state localization, we generated two distinct expressions for each object based on the ambiguous opposites and synonyms listed in Table~\ref{tab: ambiguous_opposites}. 
Overall, we obtained 25,401 expressions and 76,203 dialog instances.

\subsubsection{Evaluation Metric}
\label{sec:eva-metric}

Our main challenges are accurately predicting bounding boxes for objects in specific states and identifying grasp-relevant regions.
Following prior studies~\cite{zhang2023nextchat, you2024ferret, chen2023shikra, wang2024cogvlm}, we adopt the standard Intersection over Union (IoU)~\cite{rezatofighi2019generalized} as our evaluation metric. 
The standard IoU is a widely used measure in computer vision for tasks such as object detection and segmentation. 
Let $B_p \subset \mathbb{R}^2$ denote the predicted bounding box and 
$B_g \subset \mathbb{R}^2$ the corresponding ground-truth box. 
The intersection and union are defined as
\[
I = B_p \cap B_g, \qquad U = B_p \cup B_g.
\]
The standard IoU is
\[
\mathrm{IoU}(B_p, B_g) = \frac{|I|}{|U|},
\]
where $|\cdot|$ denotes the area measure. 
Standard IoU is bounded in [$0$, $1$].

\section{Proposed Method: StateVLM}
\label{sec:statevlm}

\subsection{Overall Structure}

We propose StateVLM, a state-aware vision-language model for fine-grained object representation learning, including object localization, object state grounding, and grasp-relevant region prediction.
We hypothesize that Pix2Seq-based approaches implicitly couple discrete text tokens with continuous spatial coordinates, leading to suboptimal bounding-box regression optimization.
Motivated by limitations in Pix2Emb-based method NExT-Chat~\cite{zhang2023nextchat}, we introduce a lightweight alternative with an auxiliary regression head that provides continuous spatial supervision during training.

During training, StateVLM jointly learns language generation and object localization from paired image-text inputs.
Given an image $I$ and a text prompt $T$, StateVLM generates a sequence of tokens $Y$ that includes both text and numerical tokens, together with the auxiliary predicted bounding box $\hat{B}$: \[ Y, \hat{B} = \texttt{StateVLM}(I, T). \]
During inference, StateVLM follows the standard Pix2Seq-style autoregressive decoding process and generates the output token sequence $Y$:
\[ Y = \texttt{StateVLM}(I, T), \] where localization information is expressed through the generated token sequence without requiring the auxiliary regression head. 
Overall, StateVLM comprises a backbone, a language modeling head (LM head), and an auxiliary regression head (AR head), as shown in Fig.~\ref{fig:method}.

\subsection{StateVLM Modules}

\subsubsection{BackBone}

\mbox{MiniCPM-V (8B)} is a compressed model that uses quantization to achieve multimodal performance comparable to large-scale models such as GPT-4V~\cite{openai2024gpt4}, while providing strong commonsense reasoning capabilities required for robotic tasks. These characteristics make MiniCPM-V an ideal testbed for our study. Therefore, we adopt MiniCPM-V~\cite{yao2025efficient} as the backbone network.

MiniCPM-V comprises three key modules: a visual decoder, a shared compression layer, and an LLM. The visual decoder utilizes an adaptive visual encoding approach, SigLip-400M~\cite{zhai2023sigmoid}, to tokenize the image inputs into visual tokens. The shared compression layer uses a perceiver-resampler structure~\cite{bai2023qwenvl} with a single cross-attention layer to compress the large number of visual tokens into a fixed set of 96 tokens. 
Finally, the compressed visual tokens and the text input are fed into the LLM (Qwen2-7B~\cite{team2024qwen2}). Therefore the backbone encoder can be formulated as a function that takes the image $I$ and text input $T$ and produces hidden representations $H$: 
\[ H = \texttt{Qwen} (\texttt{PerceiverResampler}(\texttt{SigLIP}(I)), T) \]

\begin{figure*}[h]
\centering
\includegraphics[width=0.8\textwidth]{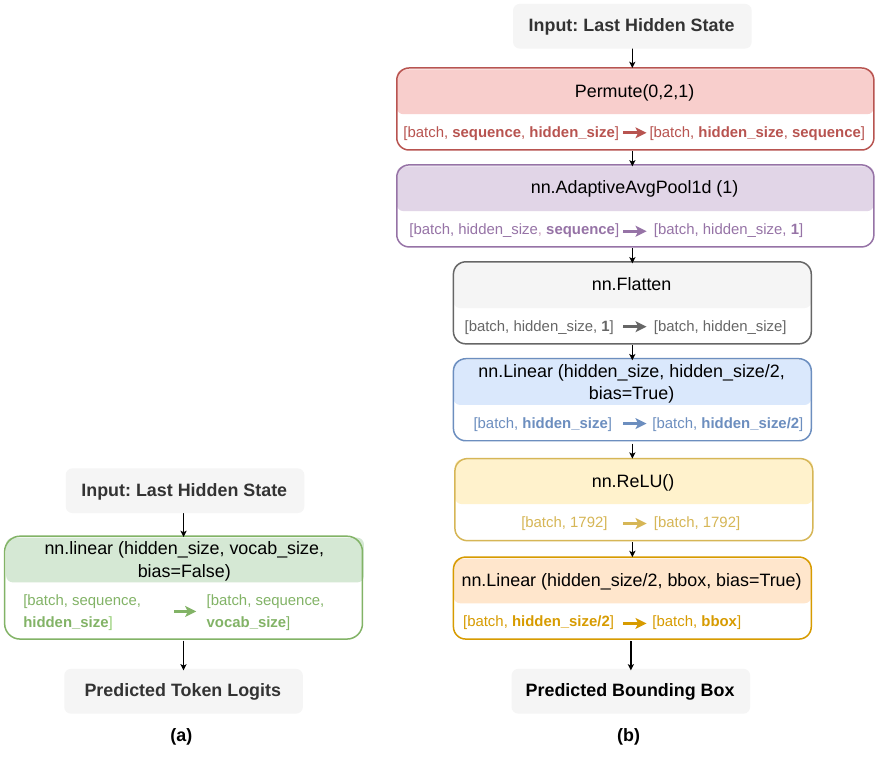}
\caption{(a) Language modeling head (LM head), The LM head corresponds to the default autoregressive training objective used in Qwen2-7B and is utilized for sequence generation in StateVLM. (b) Auxiliary regression head (AR head), The AR head represents the proposed auxiliary training objective for StateVLM.}
\label{fig:head}
\end{figure*}

\subsubsection{Language Modeling Head}

Conditioned on the hidden representations $H$, the LM head generates a sequence of tokens ($Y = {Y_1, Y_2, \dots, Y_T}$):
\[
P(Y \mid I, T)
=
\prod_{t=1}^{T}
P(y_t \mid y_{<t}, H).
\]
To preserve the model’s original language generation capability, we retain Qwen2-7B’s default LM head~\cite{team2024qwen2}, a linear layer (see Fig.~\ref{fig:head}a) used for autoregressive next-token prediction.

\subsubsection{Auxiliary Regression Head}

In parallel, we propose an AR head to predict the bounding box coordinates $\hat{B}$ from the hidden representations $H$:
\[\hat{B} = f_{\text{ARL}}(H).\]
The AR head is a lightweight feed-forward linear head that transforms the sequence output into continuous values.
As illustrated in Fig.~\ref{fig:head}b, an \texttt{AdaptiveAvgPool1d} layer is used to perform global average pooling over the temporal (sequence) dimension, reducing each hidden-state sequence to a single vector.
The resulting vector is flattened into a one-dimensional vector using a \texttt{Flatten} layer.
Next, a fully connected layer projects this vector to a lower-dimensional intermediate representation.
\texttt{ReLU} non-linearity is applied to introduce non-linear transformations, enabling the model to capture complex relationships in the data.
Finally, another fully connected layer maps the intermediate representation to a 4-dimensional output corresponding to the four continuous numbers, which are the predicted location coordinates [$x_1, y_1, x_2, y_2$].

\subsection{Joint Training Objective}
\label{sec:training-objectives}

\subsubsection{Overall Training Objective}

The overall training objective of StateVLM is a weighted combination of the \textit{causal language modeling (CLM)} loss and \textit{auxiliary regression loss (ARL)}:
\[
\mathcal{L}_{\text{CLM+ARL}} = \alpha \mathcal{L}_{\text{CLM}} + \beta \mathcal{L}_{\text{ARL}},
\]
$\alpha =0.2$ and $\beta = 0.8$ are the weights of $\mathcal{L}_{\text{CLM}}$ and $\mathcal{L}_{\text{ARL}}$, which balance these two losses and ensure they play equivalent feedback roles in model tuning. For the value selection, we tried different combinations of $\alpha$ and $\beta$ and found that this combination yields the best performance on the validation set.

\subsubsection{Causal Language Modeling}
The LLM backbone, Qwen2-7B, is trained autoregressively using the CLM objective, predicting each token conditioned on all previous tokens via a causal self-attention mask.  
Given a token sequence $[Y_1, \dots, Y_T]$ and vocabulary size $V$, with model logits $z_t \in \mathbb{R}^V$ at timestep $t$, the \textit{CLM loss} is
\[
\mathcal{L}_{\text{CLM}} = - \sum_{t=1}^{T-1} \log \frac{\exp(z_{t, Y_{t+1}})}{\sum_{v=1}^{V} \exp(z_{t, v})},
\]
where $z_{t,v}$ is the predicted logit for the token $v$ and $Y_{t+1}$ is the corresponding ground-truth token.

\subsubsection{Auxiliary Regression Loss}

To enhance numerical reasoning and improve bounding box prediction, we introduce an \textit{ARL}. 
\textit{ARL} jointly supervises the location and shape of predicted bounding boxes by combining \textit{Least Absolute Deviations (L1) loss}~\cite{brooks2013l1} and \textit{Generalized Intersection Over Union (GIoU) loss}~\cite{rezatofighi2019generalized}.

Let $B_p \subset \mathbb{R}^2$ denote the predicted bounding box and $B_g \subset \mathbb{R}^2$ the corresponding ground-truth box. The \textit{L1 loss} is computed as
\[
\mathcal{L}_{L1} = \|B_p - B_g \|_1.
\]
The \textit{GIoU loss} is based on the standard Intersection over Union (IoU) metric, which measures the overlap between two bounding boxes, as we discussed in Section~\ref{sec:eva-metric}.
Because IoU does not capture spatial discrepancies when the boxes do not overlap, GIoU introduces an additional penalty term that accounts for the normalized area outside the union of the two bounding boxes but within their smallest enclosing box.
Let $C$ denote the smallest enclosing box that contains both $B_p$ and $B_g$.
The GIoU is defined as
\[
\mathrm{GIoU}(B_p, B_g)
= \mathrm{IoU}(B_p, B_g)
- \frac{|C| - |B_p \cup B_g|}{|C|},
\]
\textit{GIoU loss} is calculated as
\[
\mathcal{L}_{\text{GIoU}} = 1 - \text{GIoU}(B_p, B_g).
\]

The total bounding box loss is a weighted sum of the two components:
\[
\mathcal{L}_{\text{ARL}} = \gamma \mathcal{L}_{L1} + \delta \mathcal{L}_{\text{GIoU}},
\]
$\gamma = 0.2$ and $ \delta = 0.8$ follows the ratio utilized in DETR~\cite{carion2020end} and NExT-Chat~\cite{zhang2023nextchat}.

\section{Experiments}
\label{sec:experiments}

\subsection{Experimental Overview}
In order to evaluate the proposed approach, we conduct two experiments. Experiment 1 evaluates the effectiveness of the proposed joint training objective by examining whether an auxiliary regression head trained with an auxiliary regression loss improves model performance. Experiment 2 assesses VLMs' understanding of object state awareness and further investigates methods for enhancing this capability.


\subsection{Implementation Details}


We implement StateVLM using PyTorch 2.1.2 and torchvision 0.16.2\footnote{\url{https://pytorch.org/}}, together with the HuggingFace Transformers library version 4.40.0.\footnote{\url{https://huggingface.co/docs/transformers/index}}
The model is initialized with weights from the pretrained MiniCPM-V 2.6\footnote{\url{https://github.com/QAdottech/MiniCPM-V}}. 
We perform full fine-tuning of the entire model in the first experiment on large-scale public benchmarks.
For the second experiment, we modify the model for parameter-efficient fine-tuning using Low-Rank Adaption (LoRA)~\cite{hu2022lora} on our proposed small-scale dataset.
All experiments are conducted on four NVIDIA A100 GPUs (80GB each). 
We employ DeepSpeed 0.14.5\footnote{\url{https://www.deepspeed.ai/}} to optimize distributed training and use an automatically warmed-up learning rate of 1e-6 to avoid unstable parameter updates during the early stages of optimization for both training regimes. 
Due to GPU memory constraints, the batch size is set to 24 for full fine-tuning and 48 for LoRA fine-tuning on four NVIDIA A100 GPUs, where LoRA enables larger batches through reduced memory overhead.
Following prior state-of-the-art approaches~\cite{lin2024sphinx, chen2023shikra, chen2023minigptv2, bai2023qwenvl, zhang2024llavag, pramanick2023jack, you2024ferret, zhang2024ferretv, wang2024cogvlm}, we evaluate predicted bounding boxes using the IoU metric (see Section~\ref{sec:eva-metric}) with a threshold of 0.5, reported as Acc@0.5.

\subsection{Experiment 1: Full Fine-Tuning on Adapted REC Benchmarks}


To evaluate the effectiveness of the proposed joint training objective by examining whether an auxiliary regression head trained with an auxiliary regression loss improves model performance, we conduct comparative experiments on the adapted REC benchmarks: RefCOCO, RefCOCO+~\cite{kazemzadeh2014referitgame}, and RefCOCOg~\cite{mao2016generation}. 
RefCOCO is split into (107859, 10834, 5657, 5095) samples for the (train, validation, testA, testB) sets, respectively. RefCOCO+ contains (107376, 10758, 5726, 4889) samples for the corresponding splits, while RefCOCOg consists of (72369, 4896, 9602) samples for the (train, validation, test) splits, respectively.
We fine-tune our models on the training split and report performance on the corresponding validation and test sets (i.e., testA and testB for RefCOCO and RefCOCO+, and the test set for RefCOCOg).

We first assess the performance of the backbone model, MiniCPM-V, as a baseline. 
Then, we conduct two groups of full fine-tuning experiments: one in which the model is trained solely with the standard \textit{CLM} objective and another in which it is trained using the \textit{CLM} combined with the \textit{ARL} objective described in Section~\ref{sec:training-objectives}.
For the distinction, we denote them as StateVLM ($\mathcal{L}_{\text{CLM}}$) and StateVLM ($\mathcal{L}_{\text{CLM+ARL}}$), respectively.

\begin{figure*}[t]
\centering
\begin{subfigure}{1\textwidth}
    \centering
    \includegraphics[width=\linewidth]{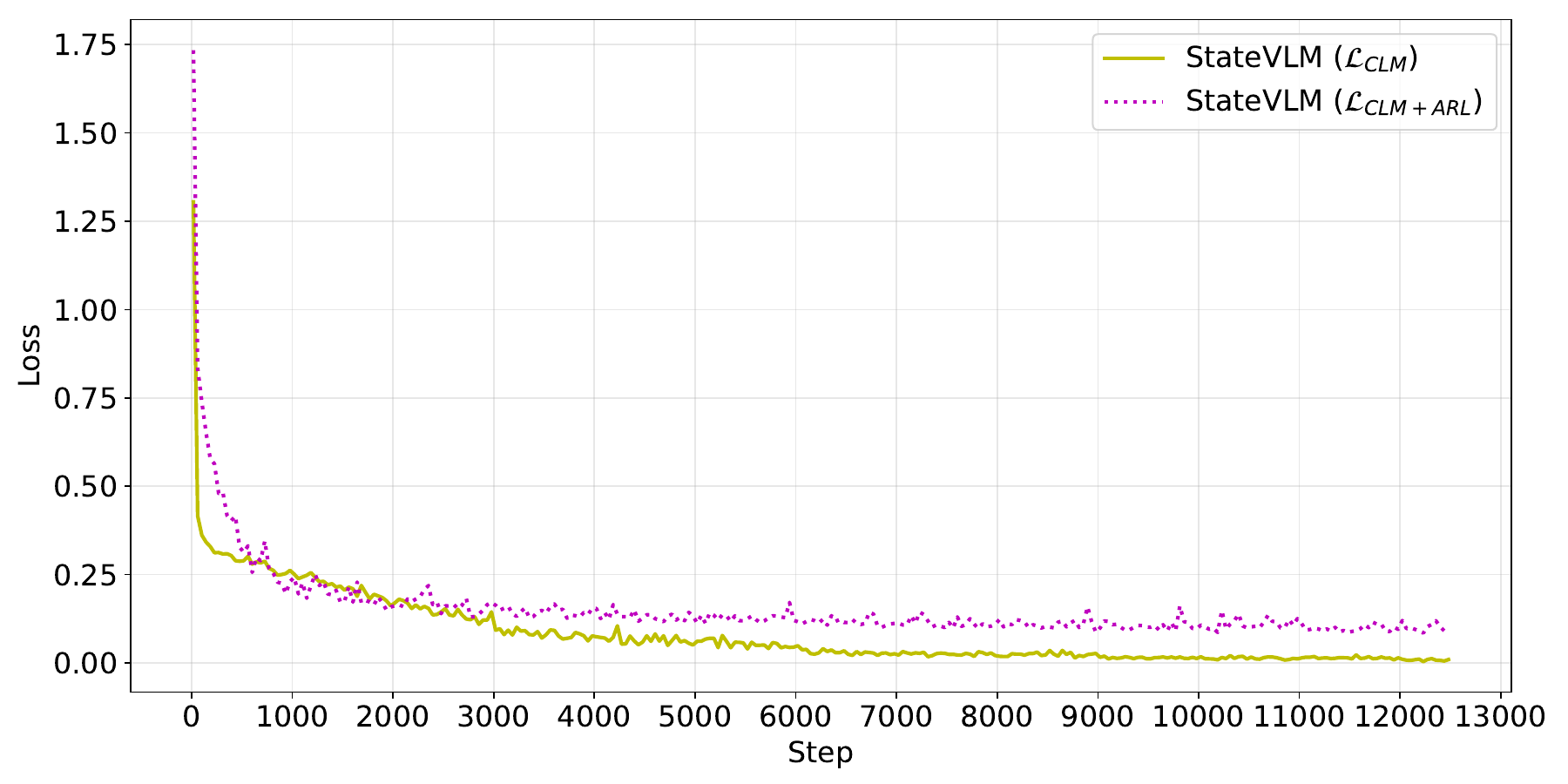}
    \caption{}
    \label{fig:loss-train}
\end{subfigure}

\vspace{0.6em}

\begin{subfigure}[t]{0.49\textwidth}
    \centering
    \includegraphics[width=\linewidth]{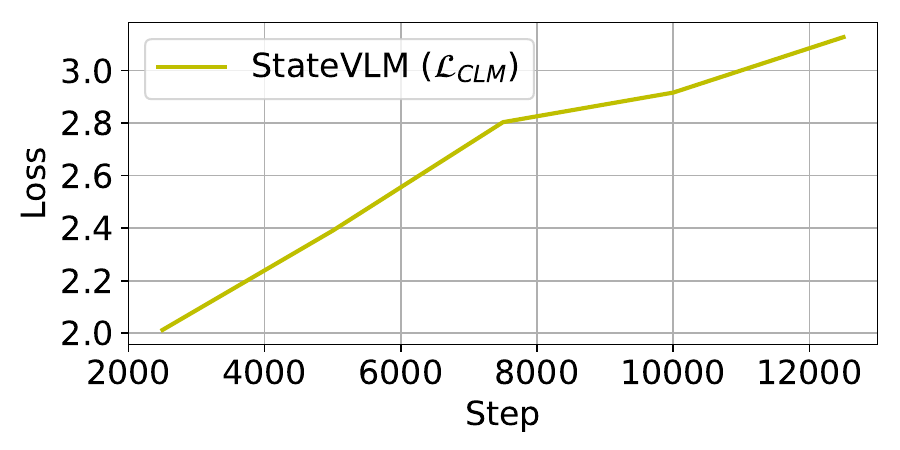}
    \caption{}
    \label{fig:loss-val1}
\end{subfigure}
\hfill
\begin{subfigure}[t]{0.49\textwidth}
    \centering
    \includegraphics[width=\linewidth]{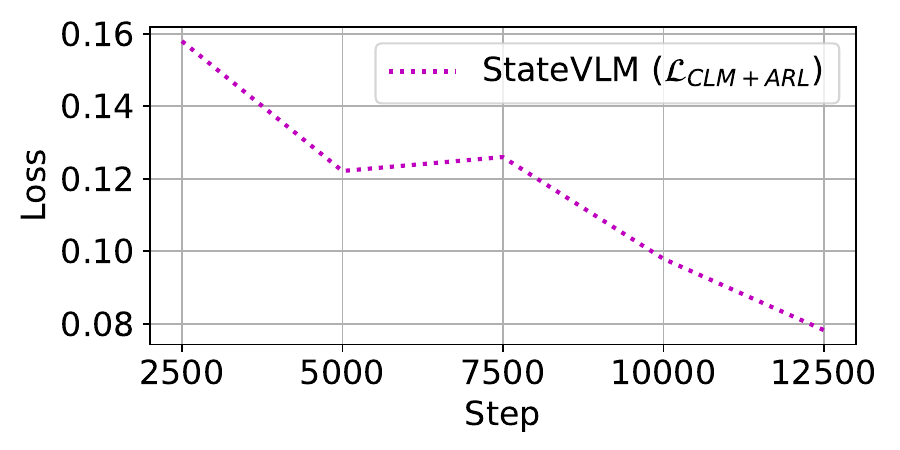}
    \caption{}
    \label{fig:loss-val2}
\end{subfigure}
\caption{StateVLM loss curves: (a) StateVLM ($\mathcal{L}_{\text{CLM}}$ and $\mathcal{L}_{\text{CLM+ARL}}$) training loss progression over steps, (b) Validation loss for \mbox{StateVLM} ($\mathcal{L}_{\text{CLM}}$), and (c) Validation loss for \mbox{StateVLM} ($\mathcal{L}_{\text{CLM+ARL}}$).}
\label{fig:loss-all}
\end{figure*}

\begin{figure*}[h]
\centering
\includegraphics[width=1\textwidth]{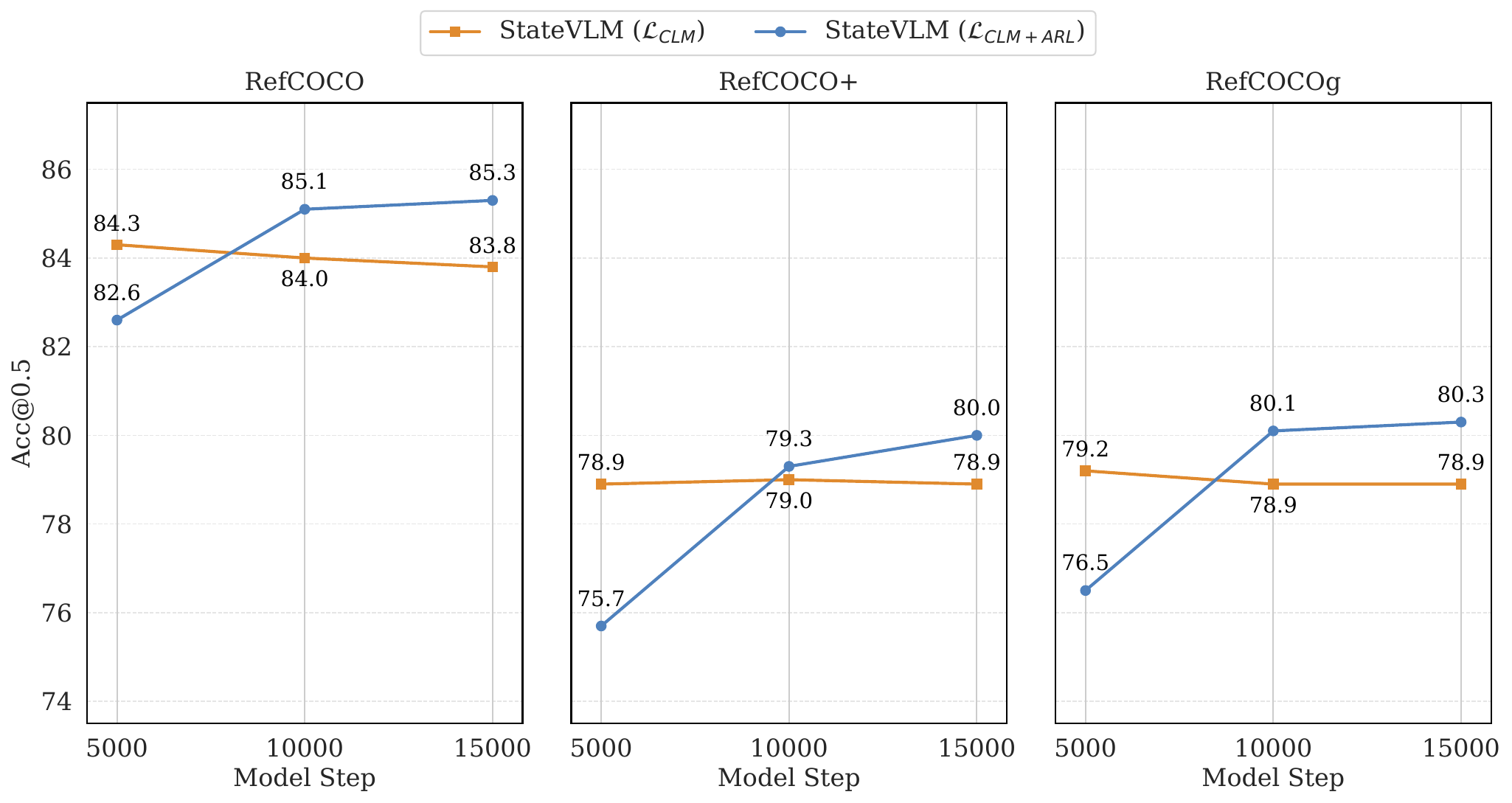}
\caption{Average performance of StateVLM ($\mathcal{L}_{\text{CLM}}$) and StateVLM ($\mathcal{L}_{\text{CLM+ARL}}$) across benchmarks over training Steps.}
\label{fig:performance}
\end{figure*}

\begin{figure*}[h]
\centering
\includegraphics[width=1\textwidth]{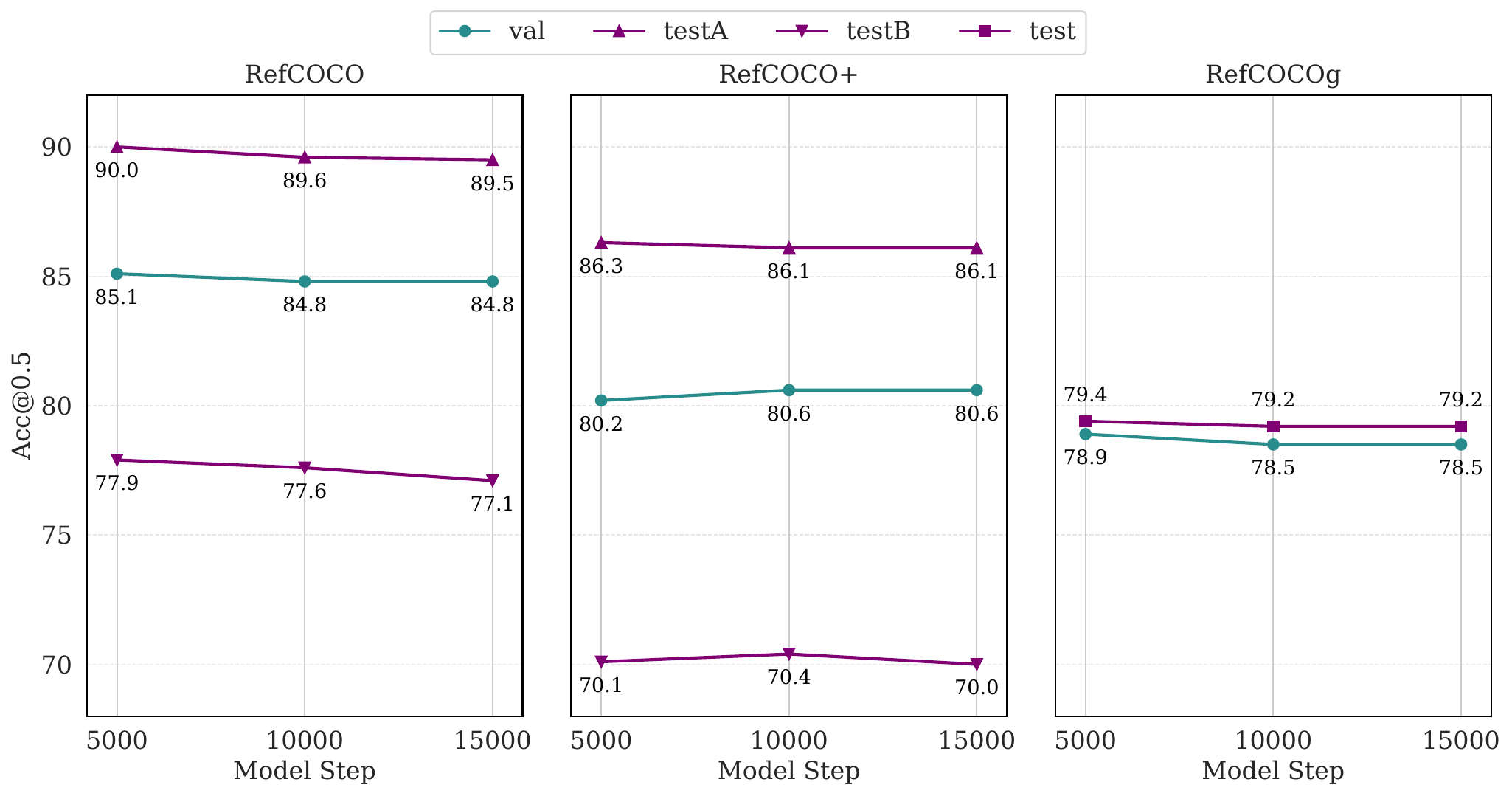}
\caption{StateVLM ($\mathcal{L}_{\text{CLM}}$) performance changes over training steps.}
\label{fig:statevlm-seq}
\end{figure*}

\begin{figure*}[h]
\centering
\includegraphics[width=1\textwidth]{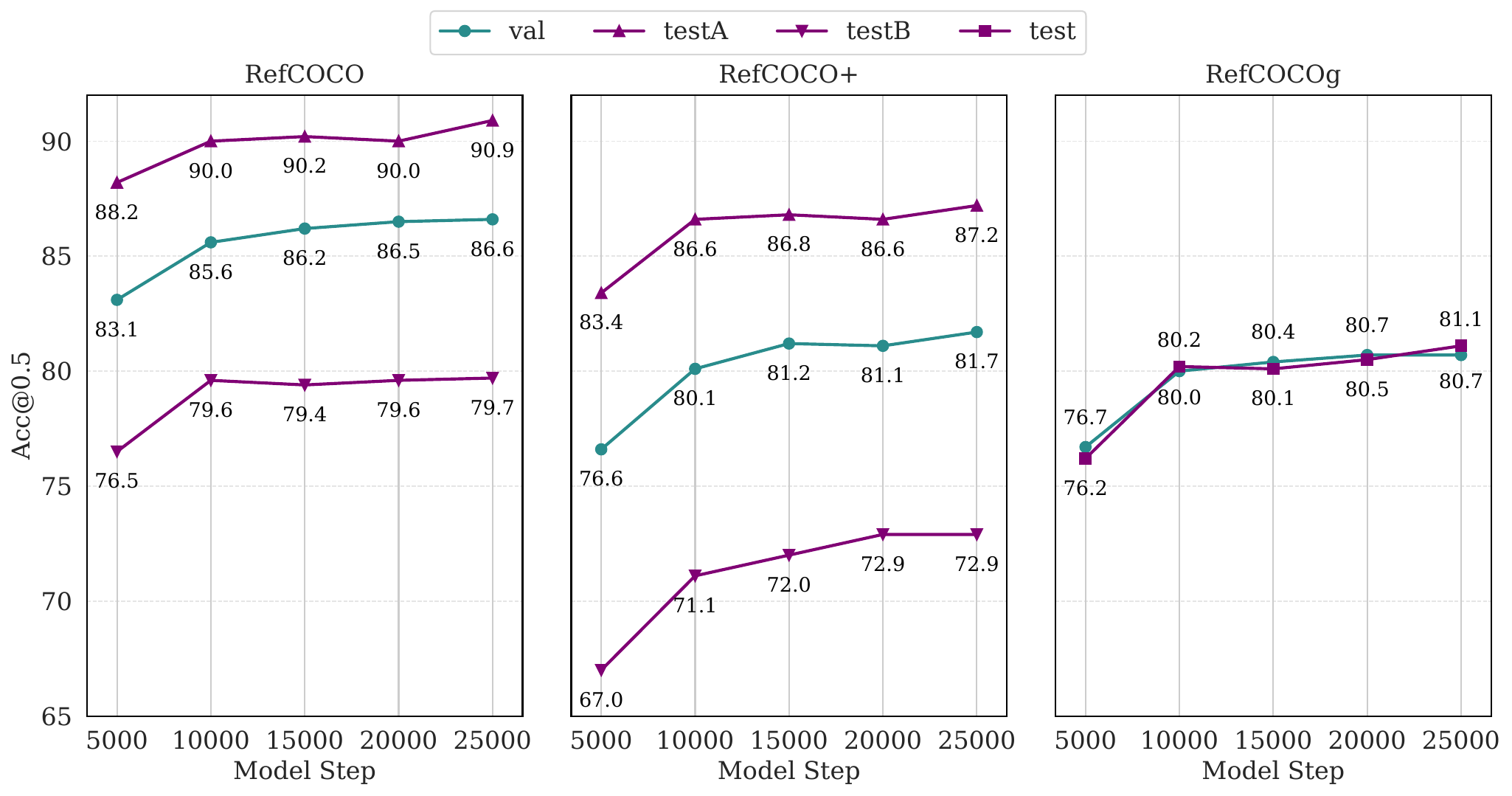}
\caption{StateVLM ($\mathcal{L}_{\text{CLM+ARL}}$) performance changes over training steps.}
\label{fig:statevlm-emb}
\end{figure*}

\subsubsection{Results and Analysis}

StateVLM ($\mathcal{L}_{\text{CLM}}$) and StateVLM ($\mathcal{L}_{\text{CLM+ARL}}$) show a similar trend, where the losses gradually decrease as the number of training steps increases, as shown in Fig.~\ref{fig:loss-train}.
The validation loss for StateVLM ($\mathcal{L}_{\text{CLM+ARL}}$) continues to decrease with the number of training steps (see Fig.~\ref{fig:loss-val2}). 
However, the validation loss for StateVLM ($\mathcal{L}_{\text{CLM}}$) continues to increase with the number of training steps (see Fig.~\ref{fig:loss-val1}). 
StateVLM ($\mathcal{L}_{\text{CLM}}$) with 5000 training steps outperforms both the 2500-step and 7500-step models.
We suspect that StateVLM ($\mathcal{L}_{\text{CLM}}$) begins to overfit after 5000 steps of training on the current training data.

Furthermore, we evaluate the fine-tuned models StateVLM ($\mathcal{L}_{\text{CLM}}$) and \mbox{StateVLM} ($\mathcal{L}_{\text{CLM+ARL}}$) on the validation and test splits of RefCOCO, RefCOCO+, and \mbox{RefCOCOg}.
We demonstrate the average performance of the two models at 5000, 10000, and 15000 training steps on the RefCOCO, RefCOCO+, and \mbox{RefCOCOg} benchmarks in Fig.~\ref{fig:performance}.
\mbox{StateVLM} ($\mathcal{L}_{\text{CLM+ARL}}$) performance improves significantly from 5000 to 15000 steps and is better than StateVLM ($\mathcal{L}_{\text{CLM}}$) after 10000 steps.
StateVLM ($\mathcal{L}_{\text{CLM}}$) performance, however, remains stable or even decreases over the training, which is consistent with the validation loss curve in Fig.~\ref{fig:loss-val1}.
The concrete performance of StateVLM ($\mathcal{L}_{\text{CLM}}$) on the subset of RefCOCO, \mbox{RefCOCO+}, and \mbox{RefCOCOg} splits is illustrated in Fig.~\ref{fig:statevlm-seq}. The performance of \mbox{StateVLM} ($\mathcal{L}_{\text{CLM}}$) remains stable between 5000 and 10000 steps and decreases at 15000 steps, indicating that it reaches its maximum performance at 5000 steps.
We continue training the StateVLM ($\mathcal{L}_{\text{CLM+ARL}}$) until 25000 steps and the performance gradually improves over the training steps on RefCOCO, RefCOCO+, and \mbox{RefCOCOg}, as shown in Fig.~\ref{fig:statevlm-emb}. 

The performance of StateVLM ($\mathcal{L}_{\text{CLM}}$) and StateVLM ($\mathcal{L}_{\text{CLM+ARL}}$) is illustrated in Table~\ref{tab:rec1}, alongside existing models on the adapted REC task.
StateVLM ($\mathcal{L}_{\text{CLM+ARL}}$) outperforms StateVLM ($\mathcal{L}_{\text{CLM}}$), demonstrating the effectiveness of the auxiliary regression loss in improving the model's performance on bounding box prediction tasks.
Compared with the Pix2Emb-based method NExT-Chat, StateVLM ($\mathcal{L}_{\text{CLM+ARL}}$) achieves superior performance.
Comparing the Pix2Seq-based method StateVLM ($\mathcal{L}_{\text{CLM}}$) with Shikra, StateVLM ($\mathcal{L}_{\text{CLM}}$) achieves lower performance than Shikra.
As the training regime, computing resources, and training time vary, we cannot directly infer why StateVLM ($\mathcal{L}_{\text{CLM}}$) underperforms Shikra.
However, StateVLM ($\mathcal{L}_{\text{CLM+ARL}}$) achieves performance comparable to Shikra, demonstrating the effectiveness of the auxiliary regression loss.
On average, the StateVLM ($\mathcal{L}_{\text{CLM+ARL}}$) achieves a 1.6\% performance improvement over the StateVLM ($\mathcal{L}_{\text{CLM}}$).
Overall, these results demonstrate that incorporating an auxiliary regression loss can significantly enhance the performance of VLMs on bounding box prediction tasks, validating our hypothesis about the limitations of the current training paradigm for object localization tasks.

\begin{table*}[t]
    \centering
    \caption{Comparative performance comparison in adapted REC benchmarks (\%). The evaluation metric is Acc@0.5.} 
    \vspace{1em}
    \label{tab:rec1}
    \resizebox{0.8\textwidth}{!}{
    \begin{tabular}{c l c c c c c c c c}
    \toprule
    \textbf{VLMs} & \multirow{2}{*}{\textbf{Methods}} & \multicolumn{3}{c}{\textbf{RefCOCO}} & \multicolumn{3}{c}{\textbf{RefCOCO+}} & \multicolumn{2}{c}{\textbf{RefCOCOg}} \\
     \cmidrule(lr){3-5} \cmidrule(lr){6-8} \cmidrule(lr){9-10} 
     \textbf{Type} & & \textbf{val} & \textbf{testA} & \textbf{testB} & \textbf{val} & \textbf{testA} & \textbf{testB} & \textbf{val} & \textbf{test} \\
    \midrule
     Pix2Emb & NExT-Chat~\cite{zhang2023nextchat} & 85.5 & 90.0 & 77.9 & 77.2 & 84.5 & 68.0 & 80.1 & 79.8 \\
    \midrule
    
    \multirow{3}{*}{Pix2Seq}
    & Shikra~\cite{chen2023shikra} & 87.0 & 90.6 & 80.2 & 81.6 & 87.4 & 72.1 & 82.3 & 82.2 \\
    \cmidrule(lr){2-10}
    & Baseline (MiniCPM-V 2.6) &  16.2 & 19.6 & 13.4 & 12.4 & 15.7 & 9.9 &  7.2 & 6.7 \\
    \cmidrule(lr){2-10}
    & \textbf{StateVLM} ($\mathcal{L}_{\text{CLM}}$) & 85.1 & 90.0 & 77.9 &  80.2 & 86.3 & 70.1 &  78.9 & 79.4 \\
    \midrule
    & \textbf{StateVLM} ($\mathcal{L}_{\text{CLM+ARL}}$) & 86.6 &  90.9 & 79.7 & 81.7 & 87.2 & 72.9 & 80.7 & 81.1\\
    \midrule
    
    & \textbf{Improvements} (Avg 1.6) & +1.5 & +0.9 & +1.8 & +1.5 & +0.9 & +2.8 & +1.8 & +1.7 \\
    \cmidrule(lr){2-10}
    \end{tabular}}
\end{table*}

\subsubsection{Ablation Studies}

We conduct ablation studies to quantify the contribution of key components in StateVLM. The auxiliary regression head is introduced to provide continuous spatial supervision. Its architectural design may influence regression accuracy. In addition, prompt formulation can affect how spatial reasoning is elicited from the language model. We further evaluate the interaction between these two factors via controlled variants.

\textit{Auxiliary Regression Head Architecture}: We study the impact of the AR head design by replacing the default two linear layers with a single linear layer. In addition, we evaluate the effect of activation functions within the two-layer variant by comparing \texttt{GELU} and \texttt{Sigmoid}. These variants isolate the influence of model depth and nonlinearity in the regression module.

\textit{Text Prompts}: We compare two prompt formulations to assess the sensitivity of StateVLM to instruction design. The first is a generic instruction prompt that provides minimal task guidance. The second is a coordinate-aware prompt that explicitly defines the image coordinate system:
\begin{itemize}
\item Generic Prompt: ``A chat between a human and an AI that understands visuals. Follow instructions.''
\item Coordinate-aware Prompt: ``A chat between a human and an AI that understands visuals. In images, [x, y] denotes points: top-left is [0, 0], bottom-right is [1, 1]. Increasing x moves right; y moves down. A bounding box is defined as [$x_1, y_1, x_2, y_2$] where ($x_1, y_1$) is the top-left corner and ($x_2, y_2$) is the bottom-right corner. Follow instructions.''
\end{itemize}

Overall, the regression head architecture shown in Fig.~\ref{fig:head}(b), when combined with the generic prompt, achieves the best performance. This result suggests that a simple regression interface, together with less constrained prompting, is sufficient for effective spatial reasoning in StateVLM under our evaluation setting.

\subsection{Experiment 2: LoRA Fine-Tuning on OSAR}

To investigate the potential object state awareness of VLMs, we introduce a small-scale and well-annotated dataset, OSAR, as detailed in Section~\ref{sec:dataset-affordance-reasoning}.
We first evaluate the performance of the baseline model, MiniCPM-V, and the models that we obtained from previous experiments: StateVLM ($\mathcal{L}_{\text{CLM}}$) and StateVLM ($\mathcal{L}_{\text{CLM+ARL}}$) on our proposed dataset.
Due to the cost of dataset generation and annotation, OSAR is relatively small-scale, which is not sufficient for full fine-tuning.
Therefore, we further fine-tune these models using the LoRA strategy, obtaining the models StateVLM ($\mathcal{L}_{\text{CLM}}$, LoRA) and StateVLM ($\mathcal{L}_{\text{CLM+ARL}}$, LoRA).

\begin{figure*}[t]
\centering
\includegraphics[width=0.7\textwidth]{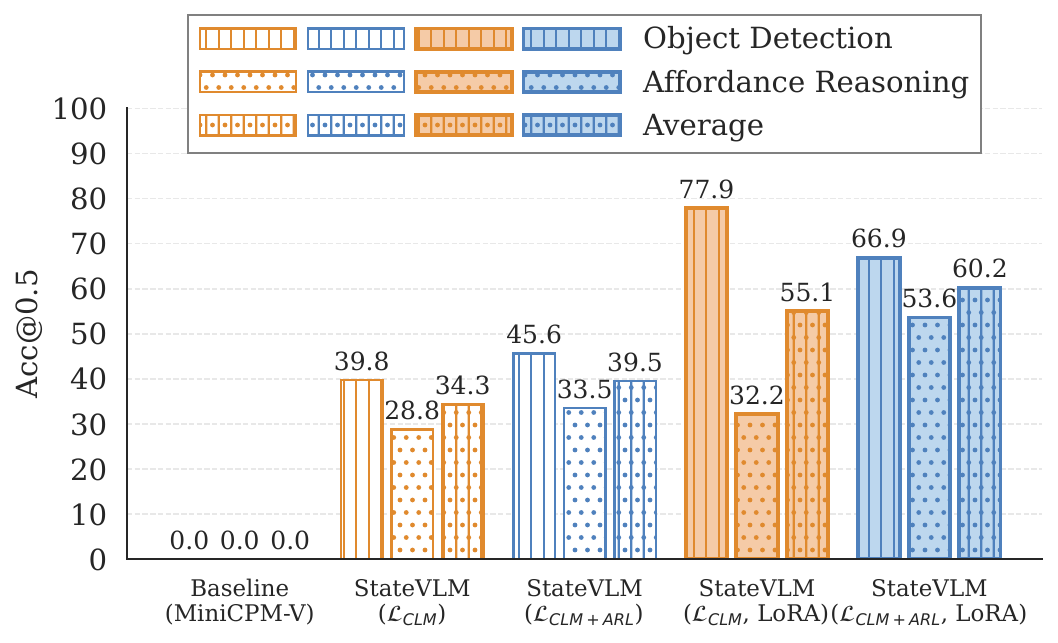}
\caption{Comprehensive performance comparison on OSAR.}
\label{fig:state-results}
\end{figure*}

\subsubsection{Results and Analysis}

A systematic experimental evaluation on the proposed tasks, object detection and affordance reasoning, is shown in Fig.~\ref{fig:state-results}.
The baseline model, MiniCPM-V 2.6, fails to perform these tasks, as it is not trained for object detection.
StateVLM ($\mathcal{L}_{\text{CLM}}$) and ($\mathcal{L}_{\text{CLM+ARL}}$), obtained from the previous experiment, achieve scores of 34.3\% and 39.5\% on OSAR, respectively. 
StateVLM ($\mathcal{L}_{\text{CLM+ARL}}$) achieves better performance than StateVLM ($\mathcal{L}_{\text{CLM}}$), demonstrating the effectiveness of the auxiliary regression loss in improving the model's performance on numerical prediction tasks. This finding is consistent with the conclusion of the first experiment.
After applying the same-pattern LoRA fine-tuning strategy on OSAR, the performance of StateVLM ($\mathcal{L}_{\text{CLM}}$, LoRA) and StateVLM ($\mathcal{L}_{\text{CLM+ARL}}$, LoRA) improve to 55.1\% and 60.2\%, respectively.
StateVLM ($\mathcal{L}_{\text{CLM+ARL}}$, LoRA) performs better than StateVLM ($\mathcal{L}_{\text{CLM}}$, LoRA).
This result further demonstrates that the auxiliary regression loss is effective in improving the model's performance on numerical prediction tasks, which is consistent with the conclusion of the first experiment.

\begin{table*}[t]
    \centering
    \caption{Exception rates(\%). N/A indicates Not Applicable because LoRA fine-tuning was evaluated only on OSAR, not on the adapted REC datasets.}
    \vspace{1em}
    \label{tab:exceptions}
    \resizebox{1\textwidth}{!}{
    \begin{tabular}{l c c c c c}
    \toprule
    \multirow{2}{*}{\textbf{Model}} & \multicolumn{3}{c}{\textbf{Adapted REC Datasets}} & \multicolumn{2}{c}{\textbf{Proprietary Dataset}} \\
    \cmidrule(lr){2-4} \cmidrule(lr){5-6}
    & \textbf{RefCOCO} & \textbf{RefCOCO+} & \textbf{RefCOCOg} & \textbf{Object Detection} & \textbf{Affordance Reasoning} \\
    \midrule
    StateVLM ($\mathcal{L}_{\text{CLM}}$) & 0.21 & 0.15 & 0.17 & 0.10 & 0.06 \\
    StateVLM ($\mathcal{L}_{\text{CLM+ARL}}$) & 0.00 & 0.00 & 0.00 & 0.00 & 0.00 \\
    StateVLM ($\mathcal{L}_{\text{CLM}}$, LoRA) & N/A & N/A & N/A & 3.45 & 33.36 \\
    StateVLM ($\mathcal{L}_{\text{CLM+ARL}}$, LoRA) & N/A & N/A & N/A & 0.00 & 0.00 \\
    \bottomrule
    \end{tabular}}
\end{table*}

Notably, StateVLM ($\mathcal{L}{\text{CLM+ARL}}$, LoRA) performs slightly worse than StateVLM ($\mathcal{L}{\text{CLM}}$, LoRA) on the object detection task, which differs from the trend observed in the first experiment. However, it consistently outperforms StateVLM ($\mathcal{L}_{\text{CLM}}$, LoRA) on the affordance reasoning task, further supporting the conclusion that the auxiliary regression loss improves performance on numerical prediction tasks.

The affordance reasoning task is more complex and challenging than the object detection task.
The object detection task only requires the model to predict the bounding box of the object, whereas the affordance reasoning task requires the model to predict the bounding box of the grasping area, which is sometimes only a part of the object.
In previous full fine-tuning benchmarks (RefCOCO, RefCOCO+, and \mbox{RefCOCOg}), the bounding boxes generally enclose the whole object rather than just a part of it.
Therefore, StateVLM ($\mathcal{L}_{\text{CLM}}$, LoRA) quickly learns to predict the bounding box of the whole object, improving from 39.8\% to 77.9\% after LoRA fine-tuning. 
However, StateVLM ($\mathcal{L}_{\text{CLM}}$, LoRA) is limited to learn to predict the bounding box of the grasping area, improving only from 28.8\% to 32.2\% after LoRA fine-tuning. StateVLM ($\mathcal{L}_{\text{CLM}}$, LoRA) also exhibits a high exception rate of 33.36\% on the affordance reasoning task, indicating that the model does not fully understand the complexity of the task and cannot predict bounding boxes in a consistent format.

In contrast, StateVLM ($\mathcal{L}_{\text{CLM+ARL}}$, LoRA) is able to learn to predict both the bounding box of the whole object and the grasping area simultaneously. It achieves a zero exception rate on both tasks, indicating that the model learns to understand the tasks and to produce bounding boxes in a consistent format.
The exception rates of all models on the subtasks are shown in Table~\ref{tab:exceptions}. Models trained with the auxiliary regression loss achieve zero exception rates, further demonstrating its effectiveness in improving model performance and output consistency on numerical tasks.
We conclude that the auxiliary regression loss enhances the model’s ability to predict bounding boxes in a consistent, predefined format, which is particularly beneficial for the affordance reasoning task.

\begin{figure*}[t]
\centering
\includegraphics[width=1\textwidth]{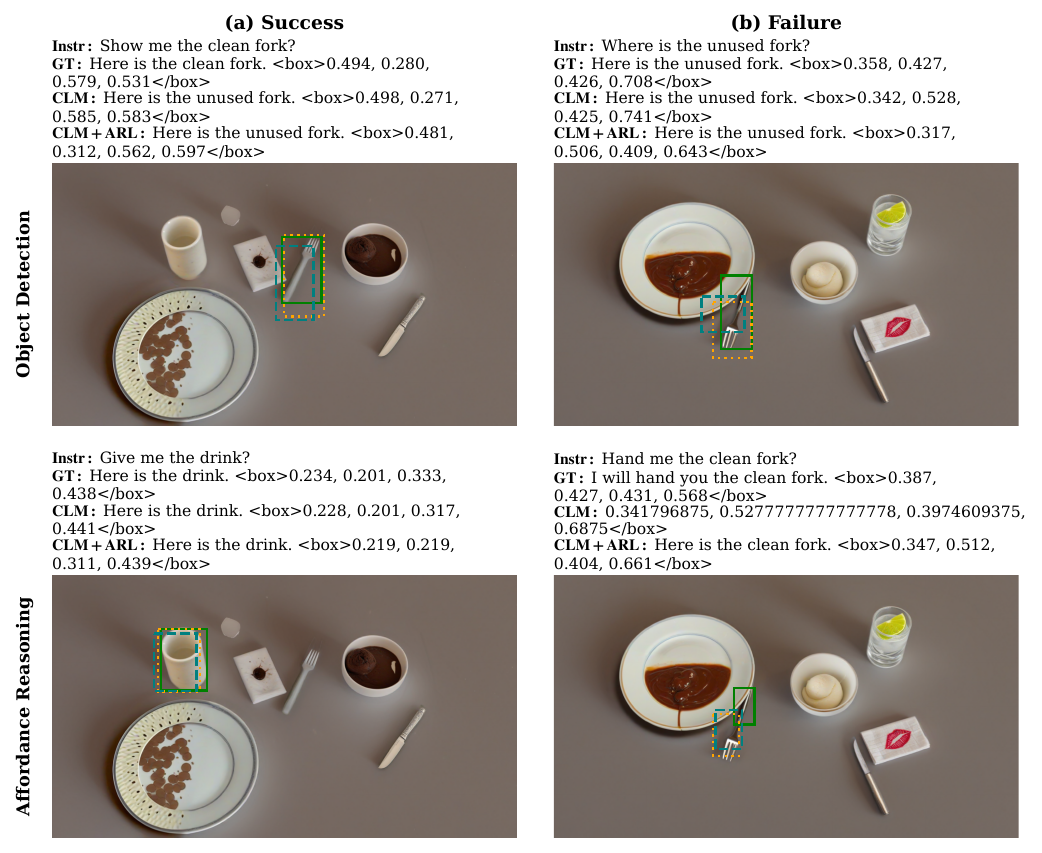}
\caption{Qualitative comparison of StateVLM ($\mathcal{L}_{\text{CLM}}$, LoRA) and StateVLM ($\mathcal{L}_{\text{CLM+ARL}}$, LoRA) on grounded object detection and affordance reasoning examples. 
Ground-truth boxes are shown in green, while model predictions are shown in orange (StateVLM ($\mathcal{L}_{\text{CLM}}$, LoRA)) and teal (StateVLM ($\mathcal{L}_{\text{CLM+ARL}}$, LoRA))).}
\label{fig:example-results}
\end{figure*}

\subsubsection{Visualization}
\label{sec:visualization}
To provide a more direct and intuitive observation of the performance of StateVLM ($\mathcal{L}_{\text{CLM}}$, LoRA) and StateVLM ($\mathcal{L}_{\text{CLM+ARL}}$, LoRA), we visualize representative successful and failed instances of predicted bounding boxes from these two models on object detection and affordance reasoning tasks, as shown in Fig.~\ref{fig:example-results}.
For object detection, the goal is to predict bounding boxes covering entire objects. 
Both models generally succeed on this task but exhibit reduced accuracy on cutlery (e.g., knives, forks, and spoons).
The predicted boxes for these cases are often biased toward one side of the object. 
This suggests a potential interference from affordance reasoning, as cutlery also represents the most challenging category for that task.
For affordance reasoning, the objective is to localize grasping regions. StateVLM ($\mathcal{L}_{\text{CLM+ARL}}$, LoRA) demonstrates improved understanding of affordances for liquids and semi-solid foods (e.g., sauces and drinks), accurately predicting grasping regions for their containers. 
In contrast, StateVLM ($\mathcal{L}_{\text{CLM}}$, LoRA) frequently produces invalid prediction formats or does not capture the intended task.
For higher-level, human-like reasoning, both models struggle to localize the appropriate regions correctly. 
For example, for a dirty plate, the desired grasp region is a clean area that allows the robotic hand to hold the plate without becoming soiled.
For cutlery, the appropriate grasp region is the handle, consistent with its intended design and function.
Overall, incorporating the auxiliary regression loss improves affordance reasoning while a significant gap remains between current performance and practical deployment, motivating further research.

\section{Conclusion and Future Work}
\label{sec:conclusion}

In this paper, we propose StateVLM, a novel model designed to perceive and learn fine-grained object representations, including precise localization of objects and their states, as well as graspable regions.
We propose a less intrusive strategy than Pix2Emb for adapting StateVLM to numerical tasks under limited computational resources.
Specifically, during the training phase, StateVLM utilizes the output of the auxiliary regression head to compute an auxiliary regression loss (ARL) for detection training.
During the inference phase, StateVLM continues with standard sequence prediction.
Comparative results on adapted REC tasks (RefCOCO, RefCOCO+, RefCOCOg) demonstrate that this ARL improves the performance of the models on REC tasks compared to training without it.
StateVLM’s performance on the proposed OSAR benchmark indicates that its latent state-aware capabilities for object state localization and affordance reasoning can be further enhanced through LoRA tuning on a small-scale and well-annotated dataset.
Additionally, we find that this ARL also enhances the consistency of StateVLM outputs, which is particularly important for complex tasks such as affordance reasoning.

There remain some challenges to our proposed approach for real-world embodied agents.
First, affordance is a broad and multifaceted concept, and we hypothesize that more extensive, carefully annotated datasets are necessary to further advance this field.
Although we leveraged a diffusion model to generate the dataset, human verification is still necessary.
Second, our evaluation focused on distinguishing objects in different states, and we did not investigate ambiguous or vague instructions in depth.
Addressing these challenges will require the development of novel approaches that extend beyond efficient fine-tuning strategies.
However, we demonstrate that incorporating an ARL improves StateVLM performance in object detection, and that a small-scale, well-annotated dataset further enhances StateVLM's state-awareness, thereby improving object state localization and affordance reasoning.

\section{Acknowledgments}

The authors gratefully acknowledge support from the China Scholarship Council (CSC) and the German Research Foundation DFG under project CML (TRR 169).
The authors also thank Tianyu Liu for his advice on the experimental design and Jae Hee Lee for his valuable feedback.

\bibliographystyle{ACM-Reference-Format}
\bibliography{bibliography}

\appendix

\section{Appendix: VLMs Configurations and Performance}\label{app1}

We summarize the configurations of VLMs capable of performing the REC task in Table~\ref{tab: vlms summary}, including their architectures, names, sizes, training resources, and training stages.
Table~\ref{tab: rec} further presents the performance of these VLMs on adapted REC benchmarks, including RefCOCO, RefCOCO+, and RefCOCOg.
Due to differences in the amount of training resources, training strategies, and data diversity, the performance of these VLMs varies significantly across benchmarks.
Our goal is not to achieve state-of-the-art performance on these benchmarks, but rather to verify the effectiveness of the auxiliary regression loss in improving VLM performance on numerical tasks, particularly object localization.

\begin{table*}[htbp]
    \centering
    \caption{Summary of VLM configurations capable of performing the REC task (\%). Pix2Seq and Pix2Emb stand for pixel-to-sequence and pixel-to-embedding. \\
                * refers to different resources to be used in the second stage. 
                $\times$ states that this stage is not included. }
    \resizebox{0.7\textwidth}{!}{
    \begin{tabular}
    {c | r | r | r | r | r}
    \toprule
    
    \textbf{VLMs} & \textbf{VLMs} & \textbf{VLMs} & \textbf{Training} & \textbf{Training} & \textbf{Training}\\
    \textbf{Architecture} &\textbf{Name} & \textbf{Size} & \textbf{GPUs} & \textbf{Stage 1} & \textbf{Stage 2}\\
    \midrule
    & Shikra~\cite{chen2023shikra} &    & 8$\times$A100 & \textasciitilde100 hours & \textasciitilde20 hours \\
    & Ferret~\cite{you2024ferret} &  & 8$\times$A100 & \textasciitilde 125 hours & $\times$  \\
    & Ferret (v2)~\cite{zhang2024ferretv} &  & 8$\times$A100 & N/A & N/A \\
    & SPHINX-1K~\cite{lin2024sphinx} & \textasciitilde13B & 32$\times$A100 & \textasciitilde 125 hours & \textasciitilde 38 hours* \\
    & SPHINX-2K~\cite{lin2024sphinx} & & 32$\times$A100& \textasciitilde 250 hours & \textasciitilde38 hours* \\
    & VistaLLM~\cite{pramanick2023jack} & & 32$\times$A100 & \textasciitilde 72 hours & \textasciitilde30 hours \\
Pix2Seq & CogVLM-G~\cite{wang2024cogvlm} & & 256$\times$A100 & \textasciitilde 120000 steps & \textasciitilde 60000 steps\\
    \cmidrule(lr){2-6} 
    & Shikra~\cite{chen2023shikra} & & 8$\times$A100 & \textasciitilde 100 hours & \textasciitilde 20 hours \\
    & MiniGPT (v2)~\cite{chen2023minigptv2} & & 8$\times$A100 & \textasciitilde 90 hours & \textasciitilde 20 hours* \\
    & Qwen-VL~\cite{bai2023qwenvl} & & N/A & \textasciitilde50000 steps & \textasciitilde19000 steps\\
    & LLaVA-G~\cite{zhang2024llavag} & \textasciitilde7B & N/A & \textasciitilde10000 steps & \textasciitilde8000 steps\\
    & VistaLLM~\cite{pramanick2023jack} & & 32$\times$A100 & \textasciitilde 48 hours & \textasciitilde22 hours \\
    & Ferret~\cite{you2024ferret} & & 8$\times$A100 & \textasciitilde 60 hours & $\times$\\
    & Ferret (v2)~\cite{zhang2024ferretv} & & 8$\times$A100 & N/A & N/A \\
    \midrule
Pix2Emb & NExT-Chat~\cite{zhang2023nextchat} & \textasciitilde7B & 8$\times$A100 & \textasciitilde59 hours & \textasciitilde10 hours\\ 
    \bottomrule
    \end{tabular}}
    \label{tab: vlms summary} 
\end{table*}

\begin{table*}[htbp]
    \centering
    \caption{StateVLM: Comparative performance comparison in REC (\%). The evaluation metric is Acc@0.5. * refers to the specialist or fine-tuned methods.} 
    \resizebox{0.8\textwidth}{!}{
    \begin{tabular}{c l c c c c c c c c}
    \toprule
    \textbf{VLMs} & \multirow{2}{*}{\textbf{Methods}} & \multicolumn{3}{c}{\textbf{RefCOCO}} & \multicolumn{3}{c}{\textbf{RefCOCO+}} & \multicolumn{2}{c}{\textbf{RefCOCOg}} \\
     \cmidrule(lr){3-5} \cmidrule(lr){6-8} \cmidrule(lr){9-10} 
     \textbf{Type} & & \textbf{val} & \textbf{testA} & \textbf{testB} & \textbf{val} & \textbf{testA} & \textbf{testB} & \textbf{val} & \textbf{test} \\
    \midrule
    \multirow{7}{*}{non-VLM}
    & MAttNet*~\cite{yu2018mattnet} & 76.4 & 80.4 & 69.3 & 64.9 & 70.3 & 56.0 & 66.7 & 67.0 \\
    & OFA-L~\cite{wang2022ofa} & 80.0 & 83.7 & 76.4 & 68.3 & 76.0 & 61.8 & 67.6 & 67.6  \\
    & TransVG*~\cite{deng2021transvg} & 81.0 & 82.7 & 78.4 & 64.8 & 70.7 & 56.9 & 68.7 & 67.7 \\
    & UNITER*~\cite{chen2020uniter} &  81.4 & 87.0 & 74.2 & 75.9 & 81.5 & 66.7 & 74.0 & 68.7 \\
    & VILLA*~\cite{gan2020villa} &  82.4 & 87.5 & 74.8 & 76.2 & 81.5 & 66.8 & 76.2 & 76.7 \\
    & UniTAB*~\cite{yang2022unitab} &  86.3 & 88.8 & 80.6 & 78.7 & 83.2 & 69.5 & 80.0 & 80.0 \\
    & G-DINO-L*~\cite{liu2023gdino} &  \textbf{90.6} & \textbf{93.2} & \textbf{88.2} & \textbf{82.8} & \textbf{89.0} & \textbf{75.9} & \textbf{86.1} & \textbf{87.0} \\
   
    \midrule
    \multirow{8}{*}{Pix2Seq-13B}
    & Shikra~\cite{chen2023shikra} & 87.8 & 91.1 & 81.8 & 82.9 & 87.8 & 74.4 & 82.6 & 83.2 \\
    & Ferret~\cite{you2024ferret} & 89.5 & 92.4 & 84.4 & 82.8 & 88.1 & 75.2 & 85.8 & 86.3 \\
    & Ferret (v2)~\cite{zhang2024ferretv} & 92.6 & \textbf{95.0} & 88.9 & 87.4 & 92.1 & 81.4 & 89.4 & 90.0 \\
    & SPHINX~\cite{lin2024sphinx} & 89.2 & 91.4 & 85.1 & 82.8 & 87.3 & 76.9 & 84.9 & 83.7 \\
    & SPHINX-1K~\cite{lin2024sphinx} & 91.1 & 92.7 & 86.7 & 86.6 & 91.1 & 80.4 & 88.2 & 88.4 \\
    & SPHINX-2k~\cite{lin2024sphinx} & 91.1 & 92.9 & 87.1 & 85.5 & 90.6 & 80.5 & 88.1 & 88.7 \\
    & VistaLLM~\cite{pramanick2023jack} & 89.9 & 92.5 & 85.0 & 84.1 & 90.3 & 75.8 & 86.0 & 86.4 \\
    & CogVLM-G~\cite{wang2024cogvlm} & \textbf{92.8} & 94.8 & \textbf{89.0} & \textbf{88.7} & \textbf{93.0} & \textbf{83.4} & \textbf{89.8} & \textbf{90.8} \\
    \cmidrule(lr){2-10} 
     \multirow{7}{*}{Pix2Seq-7B}
    & Shikra~\cite{chen2023shikra} & 87.0 & 90.6 & 80.2 & 81.6 & 87.4 & 72.1 & 82.3 & 82.2 \\
    & MiniGPT (v2)~\cite{chen2023minigptv2} & 88.1 & 91.3 & 84.3 & 79.6 & 85.5 & 73.3 & 84.2 & 84.31 \\
    & Qwen-VL~\cite{bai2023qwenvl} & 88.6 & 92.3 & 84.5 & 82.8 & 88.6 & 76.8 & 86.0 & 86.3 \\
    & LLaVA-G~\cite{zhang2024llavag} & 89.2 & --- & --- & 81.7 & --- & --- & 84.8 & --- \\
    & VistaLLM~\cite{pramanick2023jack} & 88.1 & 91.5 & 83.0 & 82.9 & 89.8 & 74.8 & 83.6 & 84.4 \\
    & Ferret~\cite{you2024ferret} & 87.5 & 91.4 & 82.5 & 80.9 & 87.4 & 73.1 & 83.9 & 84.8 \\
    & Ferret (v2)~\cite{zhang2024ferretv} & \textbf{92.8} & \textbf{94.7} & \textbf{88.7} & \textbf{87.4} & \textbf{92.8} & \textbf{79.3} & \textbf{89.4} & \textbf{89.3} \\
    \midrule
    \midrule
     Pix2Emb-7B & NExT-Chat~\cite{zhang2023nextchat} & 85.5 & 90.0 & 77.9 & 77.2 & 84.5 & 68.0 & 80.1 & 79.8 \\
    \midrule
    \midrule
    \multirow{3}{*}{Pix2Seq-7B}
    & Baseline (MiniCPM-V) &  16.2 & 19.6 & 13.4 & 12.4 & 15.7 & 9.9 &  7.2 & 6.7 \\
    \cmidrule(lr){2-10}
    & \textbf{StateVLM} ($\mathcal{L}_{\text{CLM}}$) & 85.1 & 90.0 & 77.9 &  80.2 & 86.3 & 70.1 &  78.9 & 79.4 \\
    \midrule
    & \textbf{StateVLM} ($\mathcal{L}_{\text{CLM+ARL}}$) & 86.6 &  90.9 & 79.7 & 81.7 & 87.2 & 72.9 & 80.7 & 81.1\\
    \midrule
    & \textbf{Improvements} (Avg 1.6) & +1.5 & +0.9 & +1.8 & +1.5 & +0.9 & +2.8 & +1.8 & +1.7 \\
    \cmidrule(lr){2-10}
    \end{tabular}}
    \label{tab: rec}
\end{table*}

\end{document}